\documentclass{article}
\usepackage{ijcai26}

\usepackage{times}
\usepackage{soul}
\usepackage{url}
\usepackage[hidelinks]{hyperref}
\usepackage[utf8]{inputenc}
\usepackage[small]{caption}
\usepackage{graphicx}
\usepackage{amsmath}
\usepackage{amssymb}
\usepackage{amsthm}
\usepackage{booktabs}
\usepackage{algorithm}
\usepackage{algorithmic}
\usepackage{subcaption}
\usepackage{array}
\usepackage{natbib}
\usepackage[switch]{lineno}

\linenumbers

\title{R³: Replay, Reflection, and Ranking Rewards for LLM Reinforcement Learning}

\author{
    Zhizheng Jiang$^{1,*}$\and
    Kang Zhao$^{2,*}$\and
    Weikai Xu$^1$\and
    Xinkui Lin$^3$\and
    Wei Liu$^2$\and
    Jian Luan$^2$\and
    Shuo Shang$^{1,\dagger}$\and
    Peng Han$^{1,\dagger}$ \\ 
    \affiliations
    $^1$University of Electronic Science and Technology of China\\
    $^2$Unaffiliated \\
    $^3$Institute of Information Engineering, Chinese Academy of Sciences\\
    \emails
    \{gezelligheid314, zhaokang7878\}@gmail.com, 
}

\begin{document}
\nolinenumbers
\maketitle

\let\thefootnote\relax\footnotetext{$^*$Equal contribution}
\let\thefootnote\relax\footnotetext{$^\dagger$Corresponding author}
\begin{abstract}
Large reasoning models (LRMs) aim to solve diverse and complex problems through structured reasoning. Recent advances in group-based policy optimization methods have shown promise in enabling stable advantage estimation without reliance on process-level annotations. However, these methods rely on advantage gaps induced by high-quality samples within the same batch, which makes the training process fragile and inefficient when intra-group advantages collapse under challenging tasks. 
To address these problems, we propose a reinforcement learning mechanism named \emph{\textbf{R³}} that along three directions: (1) a \emph{cross-context \underline{\textbf{R}}eplay} strategy that maintains the intra-group advantage by recalling valuable examples from historical trajectories of the same query, (2) an \emph{in-context self-\underline{\textbf{R}}eflection} mechanism enabling models to refine outputs by leveraging past failures, and (3) a \emph{structural entropy \underline{\textbf{R}}anking reward}, which assigns relative rewards to truncated or failed samples by ranking responses based on token-level entropy patterns, capturing both local exploration and global stability.
We implement our method on Deepseek-R1-Distill-Qwen-1.5B and train it on the DeepscaleR-40k in the math domain. Experiments demonstrate our method achieves SoTA performance on several math benchmarks, representing significant improvements and fewer reasoning tokens over the base models. Code and model will be released.

\end{abstract}

\section{Introduction}

Recent progress in large language models (LLMs) has led to remarkable advances in solving complex reasoning problems, with models such as OpenAI's O1 \cite{OpenAIO1} and DeepSeek's R1 \cite{guo2025deepseek} demonstrating strong step-by-step reasoning capabilities. These models often adopt the chain-of-thought (CoT) prompting paradigm \cite{wei2022chain} to encourage explicit reasoning traces. To further align LLMs with high-quality reasoning behavior, numerous open-source efforts have focused on fine-tuning models via reinforcement learning (RL).
A number of recent approaches, including VAPO~\cite{yuan2025vapo} and VC-PPO~\cite{yuan2025whatspposcollapselongcot} extend the Proximal Policy Optimization (PPO) framework~\cite{schulman2017proximal}.  In parallel, methods such as DAPO~\cite{yu2025dapo}, Open RS~\cite{dang2025reinforcement}, and Oat-Zero~\cite{liu2025understandingr1zeroliketrainingcritical}, propose enhancements to the GRPO algorithm~\cite{shao2024deepseekmath}. 


\begin{figure}[t]
  \centering
  \begin{minipage}{0.5\textwidth}
    \flushright
    \includegraphics[width=\linewidth]{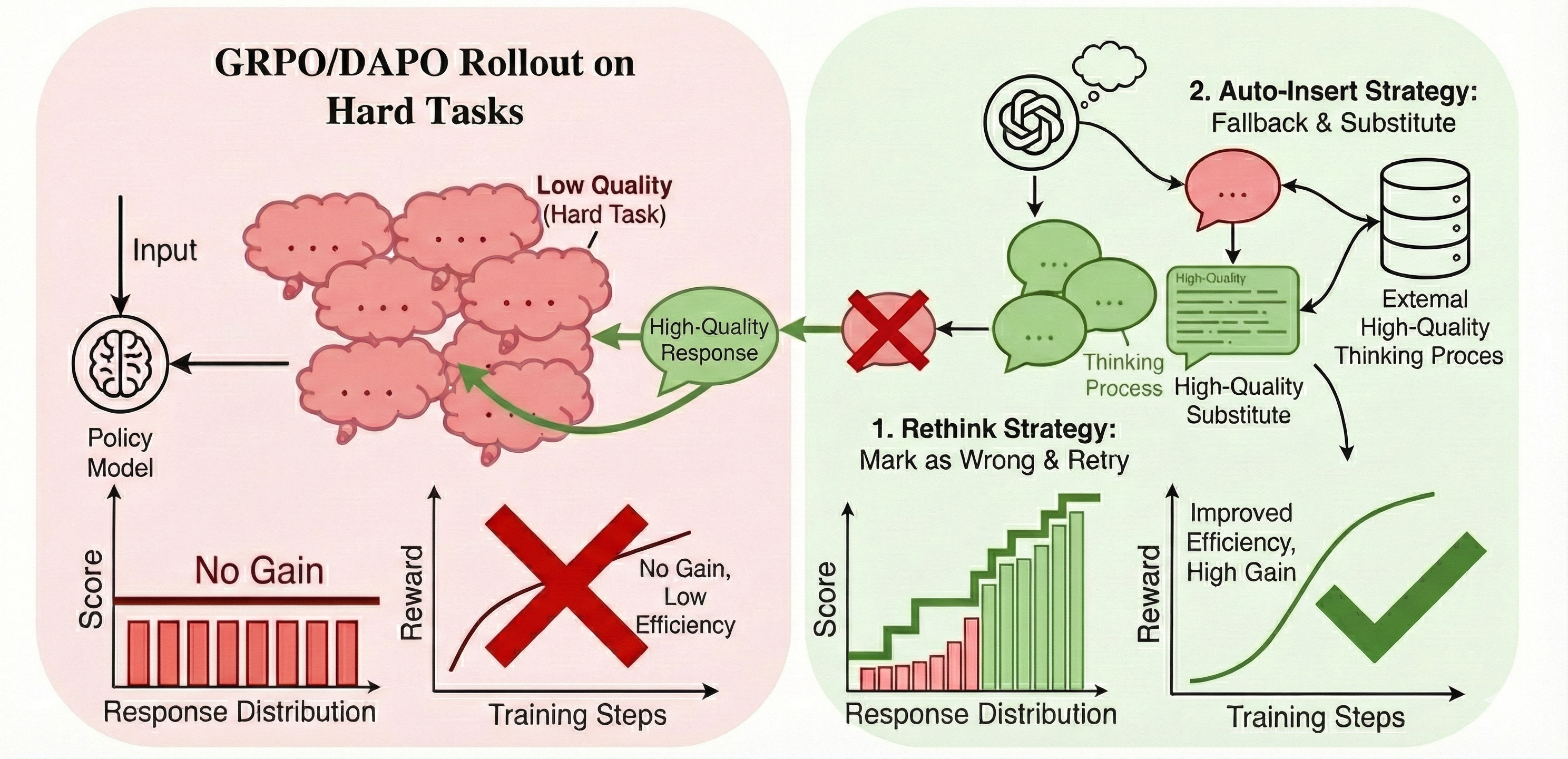}
    \caption{Multiple responses within the same group may exhibit little to no reward variation, leading to inefficient training or premature stagnation (left). Maintaining meaningful differences among intra-group samples helps preserve advantage gaps and alleviates this problem (right).
}
    \label{fig:motivation}
  \end{minipage}
\end{figure}


Although GRPO eliminates the reliance on unstable value models through intra-group advantage estimation, its use of outcome-based reward weighting further enables stable optimization of reasoning processes without requiring process reward models. However, under complex tasks, reward differences among rollout responses become inherently uncontrollable. Ideally, a more balanced reward distribution with clear relative differences within a group facilitates more accurate advantage estimation and accelerates the convergence of the policy model. In contrast, responses that are entirely incorrect, overly similar, or truncated often yield advantage collapse and provide minimal effective learning signals for training. Because a single reward cannot further differentiate these cases.

As illustrated in the left part of Figure \ref{fig:motivation}, existing methods typically update the policy model directly using the responses of a single rollout. For example, DeepscaleR \cite{luo2025deepscaler} builds upon GRPO and length-scaled reinforcement learning, a training strategy that progressively increases the response length. While DeepscaleR effectively enhances reasoning capabilities through GRPO, it heavily relies on the base competence of the policy model and lacks mechanisms to handle intra-group advantage collapse. In particular, groups dominated by low-quality responses are incorporated into training without filtering or refinement, resulting in zero or negligible positive returns on overly difficult cases. This training paradigm compromises the effectiveness of policy updates and limits further improvements.





To address these problems, we propose a reinforcement learning algorithm named \emph{\textbf{R³}} (\underline{\textbf{R}}eplay, self-\underline{\textbf{R}}eflection, and \underline{\textbf{R}}anking reward), as illustrated in the right part of Figure~\ref{fig:motivation}. R³ is designed to explicitly alleviate intra-group advantage collapse through a combination of external intervention and internal self-optimization mechanisms: 
(1) \textbf{External intervention via cross-context replay (CCR).}
To mitigate the prevalence of low-signal groups during training, we introduce an external sample buffer strategy that stores generated samples and their associated feedback over the entire training process. This buffer supports the cross-context replay mechanism, which reuses historical trajectories corresponding to the same prompt and combines them with current on-policy samples to construct mixed-policy batches. By externally injecting diverse samples, CCR ensures that each training batch maintains sufficient diversity and stable learning signals, thereby mitigating the loss of intra-group advantages and improving training stability and sample efficiency.
(2) \textbf{Internal self-optimization via in-context self-reflection (ISR).}
To reduce repetitive trial-and-error on hard queries, we introduce in-context self-reflection, which targets response batches with low reward returns. For each such sample, the buffer retrieves its prior failed attempts and integrates them with the original prompt to form a revised prompt. This design encourages the policy model to revisit its previous mistakes and refine its responses through introspective reasoning.
(3) \textbf{Structural entropy ranking reward for failed trajectories (SERR).}
To encourage effective exploration of the failure path by the policy model and further differentiate failure cases, we propose the structural entropy ranking reward. By providing relatively dense rewards, SERR distinguishes fine-grained differences among bad samples. Specifically, SERR evaluates failed trajectories based on their token-level entropy distributions, capturing the trade-off between the exploratory behavior of locally high-entropy responses and the stability of globally low-entropy ones. Crucially, by ranking these samples, SERR assigns relative rewards even to unsuccessful attempts, enabling effective learning from outputs that would otherwise be discarded.


The contributions are summarized as follows:

\begin{itemize}
    \item We propose an external replay-based intervention that preserves intra-group advantages by maintaining batch diversity.
    
    \item We design an internal mechanism that leverages failed samples via self-reflection and entropy-based ranking rewards, enabling stable advantages without process supervision.

    \item  Experiments demonstrate R³ achieves SoTA performance on several math benchmarks, representing significant improvements and fewer reasoning tokens over the base models.
\end{itemize}
\section{Related Work}
\subsection{Reinforcement Learning for Reasoning}

Recent studies aim to reproduce the reasoning capabilities of models such as OpenAI O1, whose key characteristic is explicit Chain-of-Thought (CoT) reasoning \cite{wei2022chain}. However, these approaches rely on extremely large models and large-scale RL settings.
To address this limitation, recent work explores inducing reasoning capabilities in smaller models under limited data regimes. SimpleRL \cite{c:2} demonstrates long CoT reasoning using a small number of high-quality examples and rule-based rewards. By adopting GRPO \cite{shao2024deepseekmath}, DeepScaleR shows that RL scaling effects can emerge even in small models and that iterative lengthening effectively increases reasoning trace length. Further, ProRL \cite{liu2025prorlprolongedreinforcementlearning} shows that prolonged RL with KL-divergence control \cite{kullback1951information} and reference policy resetting can uncover new reasoning behaviors beyond those of the base model.

\subsection{Experience Replay in Reinforcement Learning}

With the emergence of LRMs, experience replay (ER) \cite{mnih2015human,schaul2015prioritized,andrychowicz2017hindsight} has been revisited as a mechanism to improve sample efficiency and stabilize long-horizon reasoning. Recent works extend ER from low-level transitions to high-level reasoning trajectories. For instance, RLEP \cite{zhang2025rlep} replays verified high-quality reasoning paths to accelerate convergence and mitigate policy drift. EFrame \cite{wang2025eframe} further introduces an exploration–filter–replay loop to enable stable GRPO training.
Several methods leverage selective replay to overcome vanishing advantages and on-policy limitations. LUFFY \cite{yan2025learning} combines replayed demonstrations with on-policy rollouts to improve weak models. In multimodal and slow-thinking settings, VL-Rethinker \cite{wang2025vl} and Skywork R1V2 \cite{wang2025skywork} adopt selective replay buffers to prioritize informative trajectories.

\subsection{Entropy-Driven Exploration in Reinforcement Learning}

Entropy is a central driver of exploration in RL \cite{haarnoja2018soft,schulman2017proximal}. In LLM training, policy entropy is observed to collapse early, limiting exploration and causing premature performance saturation \cite{cui2025entropy}. Token-level analyses reveal that a small set of high-entropy “forking tokens” disproportionately influences reasoning diversity, and selectively updating these tokens improves efficiency and generalization \cite{wang2025beyond}. High-entropy regions are also linked to key reasoning behaviors such as logical pivots and self-verification \cite{cheng2025reasoning}. To mitigate entropy collapse, several entropy-aware training frameworks have been proposed \cite{zheng2025first,voelcker2025relative}. Most existing approaches, however, rely on dense supervision or discard incomplete reasoning traces. In contrast, our work introduces a fully unsupervised trajectory-level reward structural entropy ranking reward which evaluates partially correct or truncated reasoning paths based on their internal entropy structure, enabling effective learning beyond final-answer supervision.

\begin{figure*}[t]
    \centering
    \includegraphics[width=0.95\textwidth]{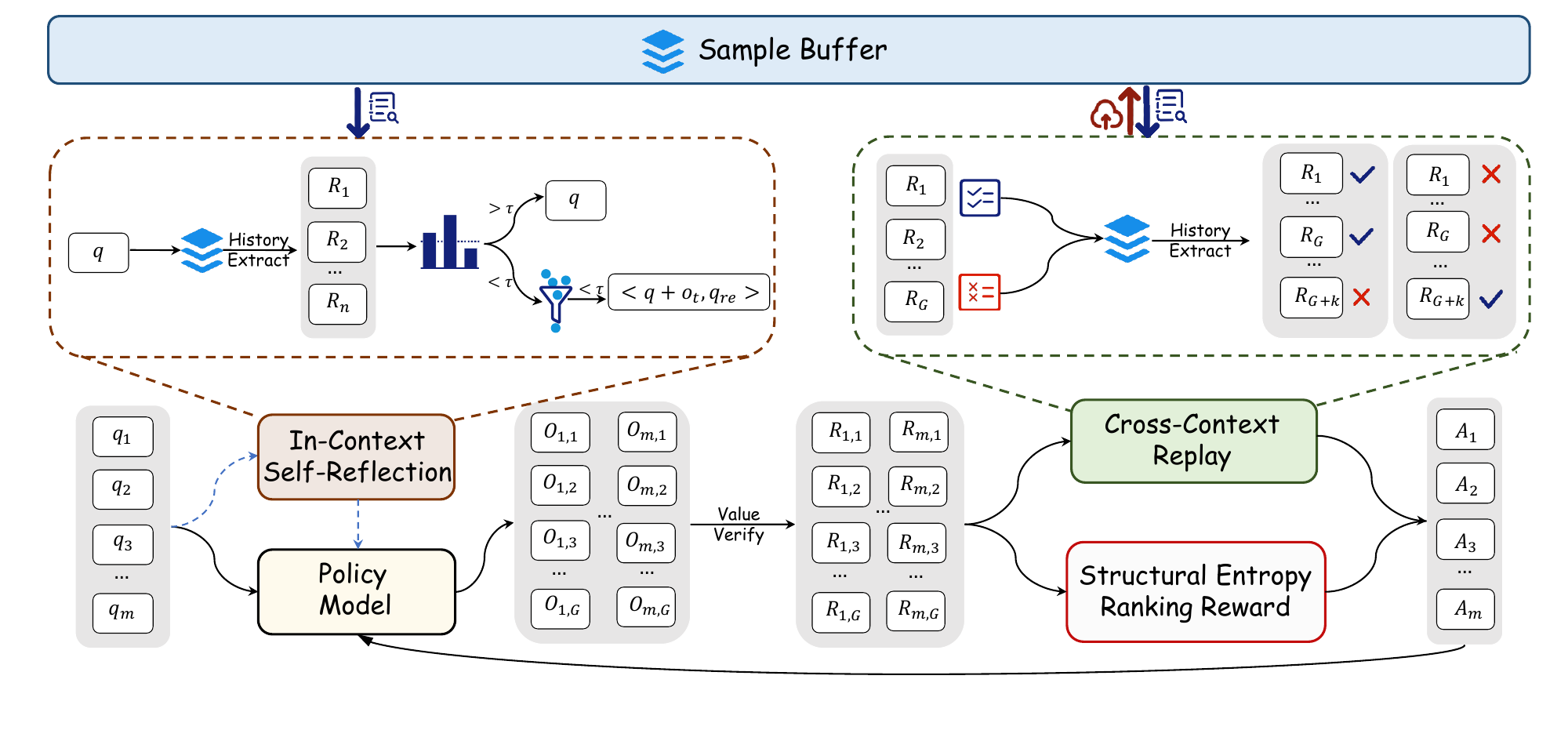}
    \caption{Overview of the R³ architecture. From left to right: (1) For hard queries from prior rounds, perform in-context self-reflection by retrieving historical samples from the sample buffer; (2) The policy model outputs a response, which is then passed through a verifier to assess its quality; (3) Cross-context replay leverages historical samples from the buffer to enhance advantage estimation; (4) Truncated responses are evaluated using the structural entropy ranking reward, which is then used for advantage computation.}
    \label{fig:Main_methodology}
\end{figure*}

\section{Methodology}

\subsection{Preliminaries: Group Relative Policy Optimization (GRPO).}
Given an input question $q$, the policy model $\pi_\theta$ generates a group of $G$ responses $\{o_i\}_{i=1}^G$ by sampling $G$ times from its output distribution. GRPO constructs a group-level reward $\{R_i\}_{i=1}^G$ from the set of sampled responses $\{o_i\}_{i=1}^G$ for the question $q$, and utilizes $\{R_i\}_{i=1}^G$  to compute the corresponding advantage for policy optimization.
\begin{equation}\label{eq:advantage}\hat{A}_{i}= \frac{R_i-\operatorname{mean}(\{R_i\}_{i=1}^G)}{\operatorname{std}(\{R_i\}_{i=1}^G)}.\end{equation}

The optimization objective of GRPO with the group-level advantage is given by the following equation:

\begin{align}
\footnotesize
\mathcal{J}_{\text{GRPO}}(\theta) = \mathbb{E}_{o \sim \pi_{\theta}} 
\Big[\frac{1}{G} \sum_{i=1}^{G} \frac{1}{|o_i|} \sum_{t=1}^{|o_i|} \min \Big(r_{i}(\theta) \hat{A_i}, \notag \\
\hspace*{\fill} \text{clip}\left(r_i(\theta), 1 - \epsilon, 1 + \epsilon\right) \hat{A_i} - {\beta}D_{KL}(\pi_{\theta}||\pi_{ref})\Big) \Big] 
\label{eq:grpo_obj}
\end{align}

where $
r_{i}(\theta) = \frac{\pi_\theta(o_{i} \mid q)}{\pi_{\theta_{\text{old}}}(o_{i} \mid q)},$ $\pi_{old}$ is the previous policy model, $\epsilon$ is the clipping coefficient and $D_{KL}$ represent computing Kullback-Leibler (KL) divergence \cite{kullback1951information} between policy model $\pi_\theta$ and reference model $\pi_{ref}$. 

While GRPO effectively stabilizes training without a critic, it fundamentally relies on intra-group reward variance to estimate advantages. As indicated in Eq.~\eqref{eq:advantage}, a critical limitation arises when all sampled responses within a group yield identical rewards. In such scenarios, the standard deviation $\operatorname{std}(\{R_i\}_{i=1}^G)$ vanishes, leading to advantage collapse. This results in uninformative gradients that hinder policy updates and impair training efficiency.

To address these challenges, we propose a stratified strategy tailored to queries of varying difficulty levels: 
(1) For medium-difficulty queries, where the model may yield homogeneous outcomes (e.g., all correct or all incorrect) within a single sampling round, relying solely on intra-group variance leads to advantage collapse. We employ \textbf{Cross-Context Replay (CCR)} to inject historical samples with opposing rewards, thereby ensuring continuous gradient updates by artificially reconstructing the advantage gap.
(2) For hard queries characterized by persistent failure, we introduce \textbf{In-Context Self-Reflection (ISR)} to explicitly guide the model toward self-correction and reasoning refinement.
(3) Finally, for extremely hard queries resulting in truncated responses where verifiable outcome rewards are unavailable, we utilize the \textbf{Structural Entropy Ranking Reward (SERR)} to provide a meaningful training signal based on the reasoning process.

\begin{algorithm*}[t]
\caption{R³ Algorithm}
\label{alg:R³}
\small
\begin{algorithmic}[1]
\REQUIRE Task Prompts $\mathcal{D}$; Policy $\pi_{\theta}$; Buffer $S$; Parameters $E, \tau, p_r$
\ENSURE Optimized Policy $\pi_\theta$
\FOR{$epoch=1:E$ \textbf{and} $\mathcal{D}_b \in \mathcal{D}$}
    \IF{$epoch > 1$}
        \STATE $\mathcal{D}_b \gets \mathcal{D}_b \cup \{ q \oplus o_h \oplus p_r \mid q \in \mathcal{D}_b, \bar{R}_q < \tau, o_h \in S \}$ \COMMENT{Self-Reflection: Augment hard queries with guided prompts}
    \ENDIF
    \STATE Sample group $G = \{o_i\}_{i=1}^{G} \sim \pi_{\theta}(\cdot \mid \mathcal{D}_b)$
    \IF{all $o_i \in G$ are negative}
        \STATE Calculate rewards $\mathcal{R}_{(k)}$ via entropy $\mathcal{E}_{\text{peak}}^{(i)}, \mathcal{E}_{\text{global}}^{(i)}$ for truncated response; $G \gets G \cup S_{pos}$ \COMMENT{Rank neg. samples via entropy; Retrieve pos. samples}
    \ELSIF{all $o_i \in G$ are positive}
        \STATE $G \gets G \cup S_{neg}$ \COMMENT{Retrieve neg. samples to form contrastive pairs}
    \ENDIF
    \STATE Update Buffer $S \gets S \cup \mathcal{D}_b$
    \STATE Update $\theta$ by maximizing R³ objective on $G$ 
\ENDFOR
\end{algorithmic}
\end{algorithm*}

\subsection{R³: Reflection, Replay, and Ranking Rewards}
The R³ framework operates on a foundation of historical experience, facilitated by a centralized sample buffer. This buffer serves as a structured repository designed to archive reasoning trajectories, providing the necessary data for cross-text learning. Each entry in the buffer comprises a query, its generated response, and the associated reward, all indexed by a unique identifier (UID) to ensure precise retrieval and traceability. Leveraging the sample buffer, R³ synergizes Cross-Context Replay (CCR) and In-Context Self-Reflection (ISR) to mitigate advantage collapse and significantly enhance sample efficiency. 
\subsubsection{Cross-Context Replay}

Inspired by the dynamic sampling strategy in DAPO \cite{yu2025dapo}, which filters out groups with 0 or 1 accuracy to ensure effective gradients, we propose a more sample-efficient alternative. Instead of discarding these zero-variance groups, we seek to actively utilize them via Cross-Context Replay (CCR). By injecting historical trajectories into these homogeneous batches, CCR restores the necessary variance for gradient estimation, ensuring that no computational resources are wasted on valid but uniform samples.

During the replay phase, we first identify queries where advantage collapse occurs due to identical rewards across all responses in group $G$. Using the unique identifier (UID) of the query, we access the sample buffer to retrieve historical trajectories associated with the same input. We specifically target samples with "opposing" rewards to construct contrastive pairs: if the current group consists entirely of negative responses (reward 0), we retrieve $k$ historical samples (reward 1); conversely, if the group is fully correct, we introduce failed attempts to re-establish a learning signal. By integrating these off-policy samples into group $G$, we achieve cross-context fusion. This explicitly restores the reward variance, ensuring the computation of a non-trivial advantage and facilitating effective training even in cases of mode collapse. 

\subsubsection{In-Context Self-Reflection} 
To mitigate persistent reasoning failures, we introduce In-Context Self-Reflection (ISR), a strategy designed to augment queries with diagnostic context derived from historical errors.

During training, for each query $q_i$ in a batch, the framework queries the sample buffer to retrieve its historical execution trajectories and corresponding rewards $\{R_{i,j}\}_{j=1}^n$. If the average historical performance $\operatorname{mean}\{R_{i,j}\}_{j=1}^n$ falls below a predefined threshold $\tau$, $q_i$ is categorized as a hard query. For such cases, we construct self-reflection exemplars by randomly sampling from the historical errors stored in the buffer. Each exemplar integrates the original query, the sampled erroneous response, and a structured reflection guidance to facilitate diagnostic reasoning. This mechanism compels the model to perform a diagnostic analysis of its past failures, thereby facilitating self-correction and preventing the recurrence of similar reasoning pitfalls in subsequent iterations.



\subsubsection{Structural Entropy Ranking Reward}
During the sampling phase of reinforcement learning, hard queries often results in truncated responses due to maximum generation length constraints. Motivated by the analysis in \cite{wang2025beyond}, which reveals that high-entropy tokens serve as "forking points" guiding divergent reasoning paths while low-entropy tokens primarily facilitate step completion, we posit that these entropy variations fundamentally reflect the trade-off between exploration and stability in reasoning. Leveraging this insight to derive reliable supervision signals for truncated solutions to challenging problems, we propose an unsupervised intrinsic reward termed \textit{structural entropy ranking reward} (SERR). Designed to provide meaningful learning signals when verifiable rewards are unavailable, SERR leverages this insight to capture the trade-off between exploration and stability. Specifically, we posit that an effective reasoning trajectory should exhibit a structural balance, where high-entropy forking points drive necessary exploration, while low-entropy tokens ensure the stability of step completion. 

We formulate two complementary entropy-based metrics to characterize the uncertainty of a generated response $o_i$ with length $L_i$.

\begin{itemize}
    \item \textbf{Peak Entropy ($\mathcal{E}_{\text{peak}}$)}: 
    Captures the intensity of divergent reasoning at critical decision steps. Aligning with the concept of "forking points," this metric quantifies the model's capacity for exploration.
    \[
    \mathcal{E}_{\text{peak}}^{(i)} = \frac{1}{|\mathcal{K}_i|} \sum_{t \in \mathcal{K}_i} H(o_{i,t}),
    \]
    where $H(o_{i,t})$ denotes the entropy of token $t$, and $\mathcal{K}_i$ represents the set of indices for the top-$k_i$ most uncertain tokens in response $o_i$. The set size is determined by $k_i = \max(1, \lfloor p \cdot L_i \rfloor)$, with $p$ being a predefined selection ratio controlling the proportion of tokens considered as critical branching points.
    
    \item \textbf{Global Entropy ($\mathcal{E}_{\text{global}}$)}: 
    Reflects the structural stability of the trajectory. Since effective reasoning relies on low-entropy tokens to facilitate consistent step completion, this metric serves as a proxy for the overall coherence of the sequence length $L_i$, ensuring that exploration remains controlled.
    \[
    \mathcal{E}_{\text{global}}^{(i)} = \frac{1}{L_i} \sum_{t=1}^{L_i} H(o_{i,t}).
    \]
    where $L_i$ is the number of non-padding tokens.
\end{itemize}

To assign a score to each sample, we define the partial order as $i \succ j$ if $\mathcal{E}_{\text{peak}}^{(i)} > \mathcal{E}_{\text{peak}}^{(j)}$ and $\mathcal{E}_{\text{global}}^{(i)} < \mathcal{E}_{\text{global}}^{(j)}$.
Each sample receives a score based on how many other samples it dominates under this partial ordering:
\[
S_i = \sum_{j \neq i} \mathbf{1}\left[i \succ j\right]
\]

Finally, we convert scores into scalar rewards by sorting all samples according to $S_i$ and assigning linearly scaled rewards from a maximum value $R_{\text{max}}$:
$$\mathcal{R}_{(k)} = R_{\text{max}} \cdot \left(1 - \frac{k}{N-1}\right)$$
where $\mathcal{R}_{(k)}$ represents the reward assigned to the sample with the $k$-th highest rank.

\subsubsection{Optimization}

The excessive integration of off-policy historical data with on-policy samples can undermine the strength of the current on-policy learning signal. Therefore, when computing the advantage for the mixed group $G_{\text{mix}}$, we apply a predefined scaling factor $\alpha$ to adjust the contribution of the off-policy samples. Based on the mixed group $G_{\text{mix}}$, the advantage is calculated as follows:
\begin{equation}\hat{A}_{i}=\frac{R_i-\operatorname{mean}(\{R_i\}_{i=1}^{G_{mix}})}{\alpha\operatorname{std}(\{R_i\}_{i=1}^{G_{mix}})+\lambda}\end{equation}
where $G_{\text{mix}} = G \cup G_C$, where $G$ denotes the on-policy generated group, and $G_C$ is the set of historical samples retrieved via the cross-context replay strategy, $\lambda$ is a small constant to ensure numerical stability.

\section{Experimental Settings}

\begin{table*}
\centering
\footnotesize
\resizebox{0.85\textwidth}{!}{
\begin{tabular}{l|ccccc|c}
\toprule
Model & AIME24 & MATH & AMC & Minerva & Olympiad & Average \\
\midrule \addlinespace[0.1cm] 
    \multicolumn{7}{c}{ \emph{\textbf{1.5B-Scale Models}}} \\
    \midrule
DeepSeek-R1-Distill-Qwen-1.5B~\citep{guo2025deepseek} & 28.12 & 82.10 & 61.21 & 26.01 & 41.59 & 47.81 \\
STILL-3-1.5B & 31.67 & 83.89 & 66.11 & 28.81 & 45.32 & 51.59 \\
DeepScaleR-1.5B-Preview~\citep{deepscaler2025} & 40.42 & 87.36 & 72.89 & 30.35 & 50.18 & 56.24 \\
FASTCURL-1.5B-V2~\citep{song2025fastcurl} &  47.50 & 89.25 & 77.01 & 32.81 & 53.28 & 59.96 \\

\midrule \addlinespace[0.1cm] 
    \multicolumn{7}{c}{ \emph{\textbf{7B-Scale Models}}} \\
    \midrule
Qwen2.5-Math-7B-Instruct & 13.34 & 79.81 & 50.62 & 34.60 & 40.69 & 43.81 \\
DeepSeek-R1-Distill-Qwen-7B & 54.16 & 91.45 & 80.19 & 37.71 & 54.55 & 63.61 \\ 
Rstar-Math-7B~\citep{guan2025rstar} & 26.74 & 78.40 & 47.51 & - & 47.11 & 49.94 \\
Eurus-2-7B-Prime~\citep{bai2025towards} & 26.74 & 79.21 & 57.84 & 38.62 & 42.10 & 48.90 \\
SimpleRL-7B~\citep{ma2025general} & 26.74 & 82.45 & 73.56 & \textbf{39.72} & 43.40 & 53.17 \\
SATURN-7B~\citep{liu2025saturn} & 48.31 & 92.65 & \textbf{85.42} & 38.96 & 55.60 & 64.20 \\
Thinker-7B\citep{chung2025thinker} & 60.00 & \textbf{92.71} & 85.09 & 38.16 & \textbf{58.62} & 66.92 \\
    
\addlinespace[0.1cm] \hline \addlinespace[0.1cm]
R³-1.5B (Ours) & \textbf{47.50} & \textbf{89.27} & \textbf{77.33} & \textbf{34.21} & \textbf{54.64} & \textbf{60.59} \\
R³-7B (Ours) & \textbf{61.04} & 92.55 & 84.18 & \underline{39.70} & \underline{58.44} & \textbf{67.18} \\

\bottomrule
\end{tabular}
}
\caption{Pass@1 performance comparison on various math benchmarks. }
\label{tab:math-results}
\end{table*}

\begin{table*}[t]
\centering
\small
\resizebox{\textwidth}{!}{
\begin{tabular}{lccccccccccccccc}
\toprule
& \multicolumn{3}{c}{AIME2024} & \multicolumn{3}{c}{MATH} & \multicolumn{3}{c}{AMC} & \multicolumn{3}{c}{Minerva} & \multicolumn{3}{c}{Olympdia} \\
\cmidrule(lr){2-4}\cmidrule(lr){5-7}\cmidrule(lr){8-10}\cmidrule(lr){11-13}\cmidrule(lr){14-16}
Model
& P@1 & P@16 & Tokens
& P@1 & P@16 & Tokens
& P@1 & P@16 & Tokens
& P@1 & P@16 & Tokens
& P@1 & P@16 & Tokens \\
\midrule
DeepSeek-Distill-1.5B & 28.1 & 70.0 & 12270.4 & 82.1 & 95.2 & 4799.6 & 61.2 & 92.7 & 8504.6 & 26.0 & 55.9 & 6299.9 & 41.5 & 62.9 & 9070.6 \\
\midrule
L1-Max & 24.6 & 60.0 & 3788.0 & 84.1 & 93.6 & 3357.6 & 65.6 & 89.2 & 3558.6 & 28.5 & 54.0 & 3414.3 & 45.4 & 61.3 & 3521.3 \\
ThinkPrune-1.5B & 23.7 & 53.3 & 5208.3 & 80.4 & 95.6 & 1652.4 &  61.3 & 90.4 & 2915.9 & 22.9 & 48.2 & 1508.7 & 40.6 & 63.7 & 3125.7 \\
Query-Opt & 26.7 & 59.8 & 10805.2 & 81.5 & 94.9 & 3284.6 & 63.2 & 90.4 & 5389.6 & 28.9 & 54.0 & 4332.8 & 47.9 & 64.1 & 5969.2 \\
O1-Pruner & 32.4 & 67.6 & 10324.4 & 71.9 & 87.5 & 3299.4 & 66.2 & 90.0 & 5993.4 & 29.2 & 56.9 & 5722.3 & 46.1 & 64.4 & 6323.2 \\
HAPO & 30.8 & 66.7 & 8967.2 & 82.8 & 95.2 & 2686.5 & 65.7 & 89.2 & 5412.8 & 25.0 & 53.3 & 2975.9 & 43.8 & 63.7 & 5892.2 \\
DeepScaleR-1.5B & 38.5 & 70.0 & 8888.0 & 87.6 & 95.6 & 3045.0 & 71.8 & 89.2 & 5529.6 & 30.9 & 54.8 & 4635.3 & 50.2 & 64.6 & 5607.6 \\
Thinker-1.5B & 27.1 & 53.3 & 4341.4 & 82.9 & 95.4 & 1829.2 & 66.2 & 91.6 & 2883.0 & 29.7 & 57.4 & 2885.3 & 46.0 & 65.9 & 3054.1 \\
FastCuRL-1.5B & 40.8 & 70.0 & 10088.7 & 87.5 & 95.6 & 3858.6 & 73.6 & 90.4 & 6682.1 & 32.1 & 58.1 & 5867.8 & 49.7 & 65.2 & 7066.7 \\
SaTurn-1.5B & 28.1 & 76.6 & 11742.6 & 82.9 & 96.0 & 4385.7 & 61.3 & 90.3 & 7928.1 & 26.7 & 51.8 & 5764.9 & 42.6 & 62.8 & 8454.3 \\
JustRL-1.5B & \textbf{53.5} & \textbf{83.4} & 9182.7 & 88.7 & 94.8 & 4097.0 & \textbf{82.2} & \textbf{93.9} & 6869.9 & 33.5 & 52.5 & 5189.1 & \textbf{55.9} & 68.8 & 7025.1 \\
\midrule
R³-1.5B (Ours) & \underline{47.5} & \underline{76.7} & 7574.1 & \textbf{89.3} & \textbf{96.0} & 3286.9 & \underline{77.3} & \underline{91.6} & 5239.9 & \textbf{34.2} & \textbf{55.9} & 4762.0 & \underline{54.6} & \textbf{69.9} & 5520.6 \\
\bottomrule
\end{tabular}
}
\caption{Comparison with different RL baseline results on DeepSeek-Distill-1.5B.}
\label{tab:main_results_1.5b}
\end{table*}
\subsection{Training}

\subsubsection{Base model.} We adopt DeepSeek-R1-Distill-Qwen-1.5B and 7B~\cite{guo2025deepseek} as base models, as they inherently possess strong reasoning capabilities, providing a solid foundation for applying our proposed training strategy R³.

\subsubsection{Datasets.} To train the model with our proposed method, we employ the DeepScaleR \cite{deepscaler2025} dataset, a high-quality synthetic dataset consisting of approximately 40,000 unique mathematics problem-answer pairs, designed to scale large language models in mathematical and logical reasoning tasks.

\subsubsection{Reward Function}
We utilize an outcome-oriented reward function based on the Qwen2.5-Math evaluator\footnote{\url{https://github.com/QwenLM/Qwen2.5-Math/tree/main/evaluation}}, which assigns correctness scores by symbolically comparing the model’s final output with the reference answer. To further promote reasoning efficiency, we incorporate a length-based reward term specifically for correct responses. Formally, given a response that matches the ground truth with length $l$ and a maximum token limit $L_{\text{max}}$ (set to 32,768), we apply an additional reward $r_{\text{len}} = \max(0, 1 - \frac{l}{L_{\text{max}}})$. This mechanism encourages the model to converge toward concise solution paths without compromising accuracy.

\subsection{Evaluation}
\subsubsection{Benchmark.} We use five widely used complex mathematical benchmarks: AIME 2024 \cite{aime2024}, MATH500 \cite{hendrycks2021measuringmathematicalproblemsolving}, AMC 2023 \cite{amc2023}, Minerva Math \cite{lewkowycz2022solvingquantitativereasoningproblems}, OlympiadBench \cite{he2024olympiadbenchchallengingbenchmarkpromoting}.
\subsubsection{Metrics.} We assess mathematical reasoning performance by generating $N=16$ independent responses per question across the benchmarks. We report Pass@1 (P@1) and Pass@16 (P@16) as evaluation metrics to capture both the standard accuracy and the model's ability to explore correct solutions within the sampling budget. Additionally, we calculate the average response length (in tokens) to analyze the inference efficiency of different models.


\section{Experiment Results}

\subsection{Overall Results}

As reported in Table~\ref{tab:math-results}, R³ achieves state-of-the-art performance across all five benchmarks. R³-1.5B obtains an average score of 60.59, significantly outperforming the base DeepSeek-R1-Distill-Qwen-1.5B by 12.78 points and surpassing strong baselines like DeepScaleR-1.5B-Preview (56.24).

More strikingly, our 1.5B model surpasses several 7B-scale models, challenging standard scaling assumptions. For instance, R³-1.5B outperforms SimpleRL-7B (53.17) and Eurus-2-7B-Prime (48.90) by substantial margins. On the challenging AIME24 benchmark, R³-1.5B scores 47.50, nearly doubling the performance of Eurus-2-7B-Prime (26.74) and matching the 7B-scale SATURN-7B (48.31).

Furthermore, when scaled to 7B, R³-7B demonstrates superior performance with an average score of 67.18, consistently outperforming robust baselines such as SATURN-7B and DeepSeek-R1-Distill-Qwen-7B. Notably, on the challenging AIME24 benchmark, R³-7B achieves a score of 61.04, surpassing the competitive Thinker-7B baseline (60.00). Similarly, on the rigorous Minerva dataset, it attains 39.70, exceeding both SATURN-7B and Thinker-7B. 

Table~\ref{tab:main_results_1.5b} presents a detailed comparison of R³-1.5B against a wide array of state-of-the-art RL baselines at the 1.5B scale. R³ achieves high accuracy with exceptional reasoning efficiency. For instance, on AIME24, the base DeepSeek-Distill-1.5B consumes 12,270 tokens to score 28.1. In contrast, R³-1.5B achieves a significantly higher score of 47.5 using only 7,574 tokens. This confirms that our framework steers the model toward concise, valid reasoning rather than redundant verification. 

\subsection{Ablation Study}
\begin{table}
\centering
\small
\resizebox{0.48\textwidth}{!}{
\begin{tabular}{l|ccccc}
\toprule
Model & AIME24 & MATH & AMC & Minerva & Olympiad  \\
\midrule \addlinespace[0.1cm] 
R³-1.5B & \textbf{47.50} & \textbf{89.27} & \textbf{77.33} & \textbf{34.21} & \textbf{54.64} \\
\addlinespace[0.1cm] \hline \addlinespace[0.1cm]
w/o CCR & 35.62  & 86.57   & 71.76    & 30.95        & 45.68         \\
w/o ISR    & 41.25  & 87.54   & 73.87   & 32.38        & 51.72        \\
w/o SERR          & 36.88  & 86.84   & 71.83   & 30.58        & 48.47         \\

\bottomrule
\end{tabular}
}

\caption{Ablation study result.}
\label{tab:ablation-results}
\end{table}

To evaluate the contribution of each component in our reinforcement learning framework, we perform ablation studies by selectively removing three key modules: cross-context replay (CCR), in-context self-reflection (ISR), and the structural entropy ranking reward (SERR). The results are summarized in Table~\ref{tab:math-results}.

Removing the CCR module leads to a clear performance drop across all benchmarks, with the most substantial declines observed on AIME24 (from 47.50 to 35.62) and Minerva (from 34.21 to 30.95). This underscores the importance of leveraging cross-query context to enhance sample efficiency and generalization. Ablating ISR also results in consistent degradation, notably on Olympiad (54.64 to 51.72) and AMC (77.33 to 73.87), indicating its effectiveness in enabling the model to refine its reasoning within a single trajectory. 
Lastly, removing SERR yields comparable drops, especially on AIME24 and Minerva, suggesting that the provision of unsupervised reward signals is crucial when standard feedback is unavailable due to truncation or failure.


\begin{figure*}[t] 
    \centering
    \includegraphics[width=0.95\textwidth]{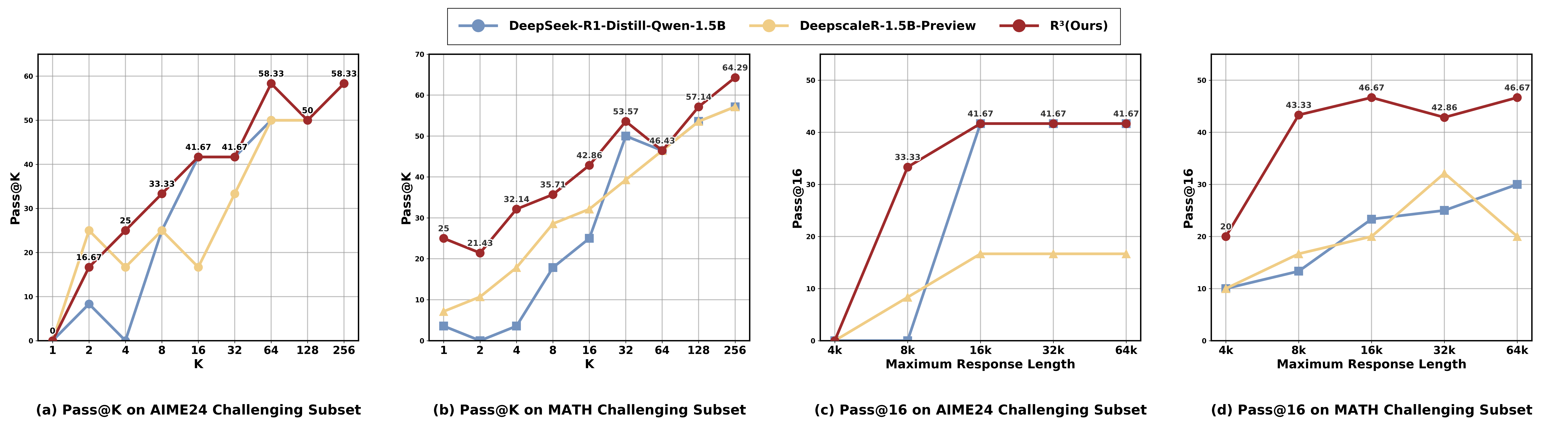} 
    \caption{Performance evaluation on challenging subsets of AIME24 and MATH}
    \label{fig:challenging_subsets}
\end{figure*}

\begin{figure}
    \centering
    \includegraphics[width=0.9\columnwidth]{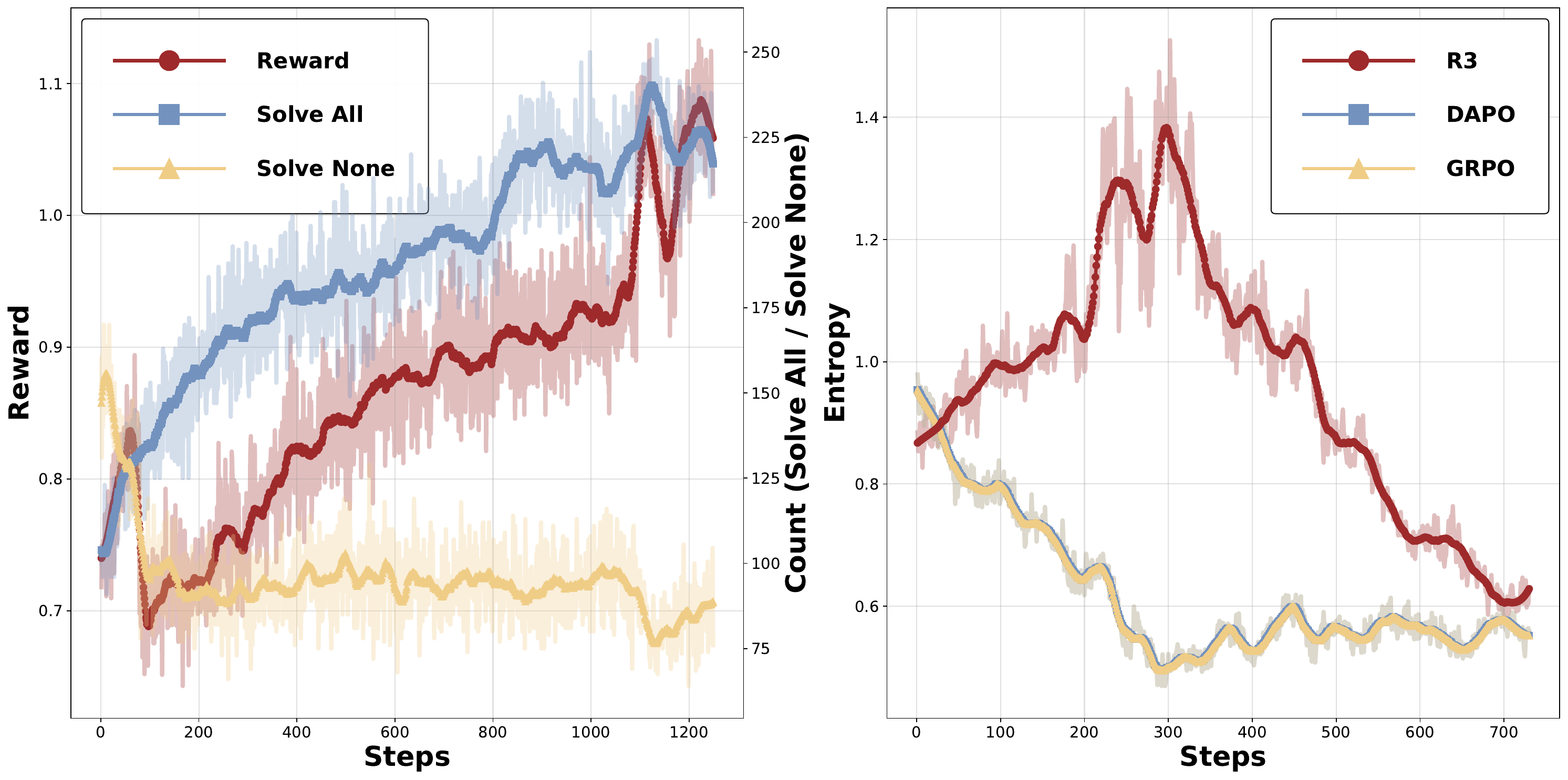}
    \caption{Training Dynamics and Entropy Evolution Analysis}
    \label{fig:training_dynamics}
\end{figure}

\subsection{Training Dynamics}

Figure~\ref{fig:training_dynamics} (left) visualizes the evolution of training metrics for R³, specifically tracking the average reward alongside the frequency of "Solve All" (all $G$ samples correct) and "Solve None" (all samples fail) groups within each sampling batch. Following an initial epoch of standard training, a notable dip in reward emerges. This shift is directly attributed to the activation of our cross-context replay (CCR) and in-context self-reflection (ISR) mechanisms. Importantly, this strategy preserves performance on mastered queries: the frequency of "Solve All" groups continues to rise steadily, while "Solve None" instances consistently decline.

Figure~\ref{fig:training_dynamics} (right) illustrates the evolution of policy entropy throughout the training process. For a rigorous comparison, we reproduced both the DAPO and GRPO baselines and plotted their corresponding entropy curves. In stark contrast to these methods, which exhibit rapid entropy decay, R³ demonstrates a distinct upward trend during the intermediate training stages. This phenomenon reflects the effective encouragement of sample exploration driven by our structural entropy ranking reward (SERR). The subsequent decline confirms that the policy eventually converges to a stable state after adequate exploration.

Taken together, these dynamics validate the synergy of our design: CCR and ISR ensure the steady mastery of complex reasoning paths (as evidenced by the "Solve All" rise), while SERR maintains the necessary exploration (via entropy retention) to prevent policy collapse.

\subsection{Experiments on Challenging Subsets}

To evaluate the effectiveness of our method on difficult tasks, we analyze model performance on challenging subsets of the AIME24 and MATH benchmarks. Specifically, we curate these subsets by selecting queries that the base model answered correctly at most once across 16 sampled responses.

We perform two sets of experiments on these challenging subsets: (1) Varying the number of sampled responses ($K$) to evaluate Pass@K, identifying the model’s upper bound in solving challenging queries (shown in Figure~\ref{fig:challenging_subsets}(a) and (b)). (2) Varying the maximum response length to evaluate Pass@16, assessing the model’s reasoning efficiency under output constraints (shown in Figure~\ref{fig:challenging_subsets}(c) and (d)).

The results on the challenging subsets of AIME24 are presented in Figure~\ref{fig:challenging_subsets}(a), respectively. We report the $\text{Pass}@K$ metric under varying numbers of sampled responses, with $K$ ranging from 1 to 256. Notably, on the challenging subset of AIME24, our model R³ begins to exhibit the ability to solve difficult queries when $K > 2$. As illustrated in Figure~\ref{fig:challenging_subsets}(b), R³ achieves the highest performance on the challenging subset of MATH with $K$ ranging from 1 to 256. Notably, while all competing approaches stagnate at lower levels, R³ surpasses this performance ceiling, achieving a $\text{Pass}@256$ score of 64.29. This indicates its effectiveness in pushing the base model beyond its inherent limitations.

As shown in Figure~\ref{fig:challenging_subsets}(c) and Figure~\ref{fig:challenging_subsets}(d), our model consistently achieves the highest scores across all tested maximum response lengths, ranging from 4k to 64k tokens. In our experiments with varying maximum response lengths, we observed that certain challenging queries could not be solved simply by increasing the length limit. This suggests that the model had reached its performance ceiling and that our R³ training enabled the base model to fully utilize its capacity.

In summary, our evaluation on challenging subsets demonstrates that R³ effectively unlocks the base model's latent reasoning capabilities. 

\section{Conclusion}
We present R³, a reinforcement learning framework that integrates Cross-Context Replay (CCR) and In-Context Self-Reflection (ISR) to stabilize training on challenging queries, while introducing the Structural Entropy Ranking Reward (SERR) to extract unsupervised signals from truncated trajectories. Experimental results demonstrate that R³ achieves state-of-the-art performance on benchmarks like AIME24, outperforming strong baselines like DeepScaleR with significantly improved sample efficiency (using only 8K generation tokens). By effectively leveraging historical and imperfect samples, R³ offers a scalable, computationally efficient pathway for advancing robust mathematical reasoning in LLMs.

\bibliographystyle{named}
\bibliography{ijcai26}

@misc{OpenAIO1,
    title = "Learning to reason with LLMs",
    author = "OpenAI",
    howpublished = "https://openai.com/index/learning-to-reason-with-llms/",
    year  = 2024,
    note  = "Accessed: 2025-4-23"   
}

@article{wei2022chain,
  title={Chain-of-thought prompting elicits reasoning in large language models},
  author={Wei, Jason and Wang, Xuezhi and Schuurmans, Dale and Bosma, Maarten and Xia, Fei and Chi, Ed and Le, Quoc V and Zhou, Denny and others},
  journal={Advances in neural information processing systems},
  volume={35},
  pages={24824--24837},
  year={2022}
}

@article{guo2025deepseek,
  title={Deepseek-r1: Incentivizing reasoning capability in llms via reinforcement learning},
  author={Guo, Daya and Yang, Dejian and Zhang, Haowei and Song, Junxiao and Zhang, Ruoyu and Xu, Runxin and Zhu, Qihao and Ma, Shirong and Wang, Peiyi and Bi, Xiao and others},
  journal={arXiv preprint arXiv:2501.12948},
  year={2025}
}

@misc{c:2,
    title = "7B Model and 8K Examples: Emerging Reasoning with Reinforcement Learning is Both Effective and Efficient",
    howpublished="https://hkust-nlp.notion.site/simplerl-reason",
    author = {Zeng, Weihao and Huang, Yuzhen and  Liu, Wei and He, Keqing and Liu, Qian and  Ma, Zejun and He, Junxian},
    year = 2025,
    note  = "Accessed: 2025-4-23" 
}

@article{luo2025deepscaler,
  title={Deepscaler: Surpassing o1-preview with a 1.5 b model by scaling rl},
  author={Luo, Michael and Tan, Sijun and Wong, Justin and Shi, Xiaoxiang and Tang, William Y and Roongta, Manan and Cai, Colin and Luo, Jeffrey and Zhang, Tianjun and Li, Li Erran and others},
  journal={Notion Blog},
  year={2025}
}

@article{shao2024deepseekmath,
  title={Deepseekmath: Pushing the limits of mathematical reasoning in open language models},
  author={Shao, Zhihong and Wang, Peiyi and Zhu, Qihao and Xu, Runxin and Song, Junxiao and Bi, Xiao and Zhang, Haowei and Zhang, Mingchuan and Li, YK and Wu, Y and others},
  journal={arXiv preprint arXiv:2402.03300},
  year={2024}
}

@article{yu2025dapo,
  title={Dapo: An open-source llm reinforcement learning system at scale},
  author={Yu, Qiying and Zhang, Zheng and Zhu, Ruofei and Yuan, Yufeng and Zuo, Xiaochen and Yue, Yu and Fan, Tiantian and Liu, Gaohong and Liu, Lingjun and Liu, Xin and others},
  journal={arXiv preprint arXiv:2503.14476},
  year={2025}
}

@article{yuan2025vapo,
  title={VAPO: Efficient and reliable reinforcement learning for advanced reasoning tasks},
  author={Yuan, Yufeng and Yu, Qiying and Zuo, Xiaochen and Zhu, Ruofei and Xu, Wenyuan and Chen, Jiaze and Wang, Chengyi and Fan, TianTian and Du, Zhengyin and Wei, Xiangpeng and others},
  journal={arXiv preprint arXiv:2504.05118},
  year={2025}
}

@misc{yuan2025whatspposcollapselongcot,
      title={What's Behind PPO's Collapse in Long-CoT? Value Optimization Holds the Secret}, 
      author={Yufeng Yuan and Yu Yue and Ruofei Zhu and Tiantian Fan and Lin Yan},
      year={2025},
      eprint={2503.01491},
      archivePrefix={arXiv},
      primaryClass={cs.LG},
      url={https://arxiv.org/abs/2503.01491}, 
}

@article{schulman2017proximal,
  title={Proximal policy optimization algorithms},
  author={Schulman, John and Wolski, Filip and Dhariwal, Prafulla and Radford, Alec and Klimov, Oleg},
  journal={arXiv preprint arXiv:1707.06347},
  year={2017}
}

@article{dang2025reinforcement,
  title={Reinforcement Learning for Reasoning in Small LLMs: What Works and What Doesn't},
  author={Dang, Quy-Anh and Ngo, Chris},
  journal={arXiv preprint arXiv:2503.16219},
  year={2025}
}

@misc{liu2025understandingr1zeroliketrainingcritical,
      title={Understanding R1-Zero-Like Training: A Critical Perspective}, 
      author={Zichen Liu and Changyu Chen and Wenjun Li and Penghui Qi and Tianyu Pang and Chao Du and Wee Sun Lee and Min Lin},
      year={2025},
      eprint={2503.20783},
      archivePrefix={arXiv},
      primaryClass={cs.LG},
      url={https://arxiv.org/abs/2503.20783}, 
}

@misc{aime2024,
 author = {{Mathematical Association of America}},
 title = {Aime problems and solutions.},
 year = 2024,
 howpublished={\url{https://artofproblemsolving.com/wiki/index.php/AIME_Problems_and_Solutions}},
 note = {Accessed: 2025-5-28}
}

@misc{hendrycks2021measuringmathematicalproblemsolving,
      title={Measuring Mathematical Problem Solving With the MATH Dataset}, 
      author={Dan Hendrycks and Collin Burns and Saurav Kadavath and Akul Arora and Steven Basart and Eric Tang and Dawn Song and Jacob Steinhardt},
      year={2021},
      eprint={2103.03874},
      archivePrefix={arXiv},
      primaryClass={cs.LG},
      url={https://arxiv.org/abs/2103.03874}, 
}

@misc{amc2023,
  author = {{Mathematical Association of America}},
  title = {Amc problems and solutions.},
  year = 2023,
  howpublished={\url{https://artofproblemsolving.com/wiki/index.php?title=AMC_Problems_and_Solutions}},
  note = {Accessed: 2025-5-28}
}

@misc{lewkowycz2022solvingquantitativereasoningproblems,
      title={Solving Quantitative Reasoning Problems with Language Models}, 
      author={Aitor Lewkowycz and Anders Andreassen and David Dohan and Ethan Dyer and Henryk Michalewski and Vinay Ramasesh and Ambrose Slone and Cem Anil and Imanol Schlag and Theo Gutman-Solo and Yuhuai Wu and Behnam Neyshabur and Guy Gur-Ari and Vedant Misra},
      year={2022},
      eprint={2206.14858},
      archivePrefix={arXiv},
      primaryClass={cs.CL},
      url={https://arxiv.org/abs/2206.14858}, 
}

@misc{he2024olympiadbenchchallengingbenchmarkpromoting,
      title={OlympiadBench: A Challenging Benchmark for Promoting AGI with Olympiad-Level Bilingual Multimodal Scientific Problems}, 
      author={Chaoqun He and Renjie Luo and Yuzhuo Bai and Shengding Hu and Zhen Leng Thai and Junhao Shen and Jinyi Hu and Xu Han and Yujie Huang and Yuxiang Zhang and Jie Liu and Lei Qi and Zhiyuan Liu and Maosong Sun},
      year={2024},
      eprint={2402.14008},
      archivePrefix={arXiv},
      primaryClass={cs.CL},
      url={https://arxiv.org/abs/2402.14008}, 
}

@misc{deepscaler2025,
  title={DeepScaleR: Surpassing O1-Preview with a 1.5B Model by Scaling RL},
  author={Michael Luo and Sijun Tan and Justin Wong and Xiaoxiang Shi and William Y. Tang and Manan Roongta and Colin Cai and Jeffrey Luo and Li Erran Li and Raluca Ada Popa and Ion Stoica},
  year={2025},
  howpublished={\url{https://pretty-radio-b75.notion.site/DeepScaleR-Surpassing-O1-Preview-with-a-1-5B-Model-by-Scaling-RL-19681902c1468005bed8ca303013a4e2}},
  note={Notion Blog},
  year={2025}
}

@misc{liu2025prorlprolongedreinforcementlearning,
      title={ProRL: Prolonged Reinforcement Learning Expands Reasoning Boundaries in Large Language Models}, 
      author={Mingjie Liu and Shizhe Diao and Ximing Lu and Jian Hu and Xin Dong and Yejin Choi and Jan Kautz and Yi Dong},
      year={2025},
      eprint={2505.24864},
      archivePrefix={arXiv},
      primaryClass={cs.CL},
      url={https://arxiv.org/abs/2505.24864}, 
}

@inproceedings{haarnoja2018soft,
  title={Soft actor-critic: Off-policy maximum entropy deep reinforcement learning with a stochastic actor},
  author={Haarnoja, Tuomas and Zhou, Aurick and Abbeel, Pieter and Levine, Sergey},
  booktitle={International conference on machine learning},
  pages={1861--1870},
  year={2018},
  organization={Pmlr}
}

@article{cui2025entropy,
  title={The entropy mechanism of reinforcement learning for reasoning language models},
  author={Cui, Ganqu and Zhang, Yuchen and Chen, Jiacheng and Yuan, Lifan and Wang, Zhi and Zuo, Yuxin and Li, Haozhan and Fan, Yuchen and Chen, Huayu and Chen, Weize and others},
  journal={arXiv preprint arXiv:2505.22617},
  year={2025}
}

@article{wang2025beyond,
  title={Beyond the 80/20 rule: High-entropy minority tokens drive effective reinforcement learning for llm reasoning},
  author={Wang, Shenzhi and Yu, Le and Gao, Chang and Zheng, Chujie and Liu, Shixuan and Lu, Rui and Dang, Kai and Chen, Xionghui and Yang, Jianxin and Zhang, Zhenru and others},
  journal={arXiv preprint arXiv:2506.01939},
  year={2025}
}

@article{cheng2025reasoning,
  title={Reasoning with exploration: An entropy perspective},
  author={Cheng, Daixuan and Huang, Shaohan and Zhu, Xuekai and Dai, Bo and Zhao, Wayne Xin and Zhang, Zhenliang and Wei, Furu},
  journal={arXiv preprint arXiv:2506.14758},
  year={2025}
}

@article{zheng2025first,
  title={First Return, Entropy-Eliciting Explore},
  author={Zheng, Tianyu and Xing, Tianshun and Gu, Qingshui and Liang, Taoran and Qu, Xingwei and Zhou, Xin and Li, Yizhi and Wen, Zhoufutu and Lin, Chenghua and Huang, Wenhao and others},
  journal={arXiv preprint arXiv:2507.07017},
  year={2025}
}

@article{voelcker2025relative,
  title={Relative Entropy Pathwise Policy Optimization},
  author={Voelcker, Claas and Brunnbauer, Axel and Hussing, Marcel and Nauman, Michal and Abbeel, Pieter and Eaton, Eric and Grosu, Radu and Farahmand, Amir-massoud and Gilitschenski, Igor},
  journal={arXiv preprint arXiv:2507.11019},
  year={2025}
}

@article{kullback1951information,
  title={On information and sufficiency},
  author={Kullback, Solomon and Leibler, Richard A},
  journal={The annals of mathematical statistics},
  volume={22},
  number={1},
  pages={79--86},
  year={1951},
  publisher={JSTOR}
}

@article{zhang2025rlep,
  title={RLEP: Reinforcement Learning with Experience Replay for LLM Reasoning},
  author={Zhang, Hongzhi and Fu, Jia and Zhang, Jingyuan and Fu, Kai and Wang, Qi and Zhang, Fuzheng and Zhou, Guorui},
  journal={arXiv preprint arXiv:2507.07451},
  year={2025}
}

@article{wang2025eframe,
  title={EFRame: Deeper Reasoning via Exploration-Filtering-Replay Reinforcement Learning Framework},
  author={Wang, Chen and Wei, Lai and Zhang, Yanzhi and Shao, Chenyang and Dan, Zedong and Huang, Weiran and Wang, Yue and Zhang, Yuzhi},
  journal={arXiv preprint arXiv:2506.22200},
  year={2025}
}

@article{yan2025learning,
  title={Learning to reason under off-policy guidance},
  author={Yan, Jianhao and Li, Yafu and Hu, Zican and Wang, Zhi and Cui, Ganqu and Qu, Xiaoye and Cheng, Yu and Zhang, Yue},
  journal={arXiv preprint arXiv:2504.14945},
  year={2025}
}

@article{wang2025vl,
  title={Vl-rethinker: Incentivizing self-reflection of vision-language models with reinforcement learning},
  author={Wang, Haozhe and Qu, Chao and Huang, Zuming and Chu, Wei and Lin, Fangzhen and Chen, Wenhu},
  journal={arXiv preprint arXiv:2504.08837},
  year={2025}
}

@article{wang2025skywork,
  title={Skywork r1v2: Multimodal hybrid reinforcement learning for reasoning},
  author={Wang, Peiyu and Wei, Yichen and Peng, Yi and Wang, Xiaokun and Qiu, Weijie and Shen, Wei and Xie, Tianyidan and Pei, Jiangbo and Zhang, Jianhao and Hao, Yunzhuo and others},
  journal={arXiv preprint arXiv:2504.16656},
  year={2025}
}

@article{mnih2015human,
  title={Human-level control through deep reinforcement learning},
  author={Mnih, Volodymyr and Kavukcuoglu, Koray and Silver, David and Rusu, Andrei A and Veness, Joel and Bellemare, Marc G and Graves, Alex and Riedmiller, Martin and Fidjeland, Andreas K and Ostrovski, Georg and others},
  journal={nature},
  volume={518},
  number={7540},
  pages={529--533},
  year={2015},
  publisher={Nature Publishing Group}
}

@article{schaul2015prioritized,
  title={Prioritized experience replay},
  author={Schaul, Tom and Quan, John and Antonoglou, Ioannis and Silver, David},
  journal={arXiv preprint arXiv:1511.05952},
  year={2015}
}

@article{andrychowicz2017hindsight,
  title={Hindsight experience replay},
  author={Andrychowicz, Marcin and Wolski, Filip and Ray, Alex and Schneider, Jonas and Fong, Rachel and Welinder, Peter and McGrew, Bob and Tobin, Josh and Pieter Abbeel, OpenAI and Zaremba, Wojciech},
  journal={Advances in neural information processing systems},
  volume={30},
  year={2017}
}

@article{liu2025saturn,
  title={SATURN: SAT-based Reinforcement Learning to Unleash Language Model Reasoning},
  author={Liu, Huanyu and Li, Jia and Zhu, Hao and Zhang, Kechi and Dong, Yihong and Li, Ge},
  journal={arXiv preprint arXiv:2505.16368},
  year={2025}
}

@article{chung2025thinker,
  title={Thinker: Learning to Think Fast and Slow},
  author={Chung, Stephen and Du, Wenyu and Fu, Jie},
  journal={arXiv preprint arXiv:2505.21097},
  year={2025}
}

@article{song2025fastcurl,
  title={FastCuRL: Curriculum Reinforcement Learning with Stage-wise Context Scaling for Efficient Training R1-like Reasoning Models},
  author={Song, Mingyang and Zheng, Mao and Li, Zheng and Yang, Wenjie and Luo, Xuan and Pan, Yue and Zhang, Feng},
  journal={arXiv preprint arXiv:2503.17287},
  year={2025}
}

@article{guan2025rstar,
  title={rStar-Math: Small LLMs Can Master Math Reasoning with Self-Evolved Deep Thinking},
  author={Guan, Xinyu and Zhang, Li Lyna and Liu, Yifei and Shang, Ning and Sun, Youran and Zhu, Yi and Yang, Fan and Yang, Mao},
  journal={arXiv preprint arXiv:2501.04519},
  year={2025}
}

@article{bai2025towards,
  title={Towards Effective Code-Integrated Reasoning},
  author={Bai, Fei and Min, Yingqian and Zhang, Beichen and Chen, Zhipeng and Zhao, Wayne Xin and Fang, Lei and Liu, Zheng and Wang, Zhongyuan and Wen, Ji-Rong},
  journal={arXiv preprint arXiv:2505.24480},
  year={2025}
}

@article{ma2025general,
  title={General-reasoner: Advancing llm reasoning across all domains},
  author={Ma, Xueguang and Liu, Qian and Jiang, Dongfu and Zhang, Ge and Ma, Zejun and Chen, Wenhu},
  journal={arXiv preprint arXiv:2505.14652},
  year={2025}
}

\end{document}